# Enhanced Genetic Algorithm approach for Solving Dynamic Shortest Path Routing Problems using Immigrants and Memory Schemes

Dr. T. R. Gopalakrishnan Nair , Ms.Kavitha Sooda, Ms.Yashoda M B

*Abstract*—In Internet Routing, the static shortest path (SP) problem has been addressed using well known intelligent optimization techniques like artificial neural networks, genetic algorithms (GAs) and particle swarm optimization. Advancement in wireless communication lead more and more mobile wireless networks, such as mobile networks [mobile ad hoc networks (MANETs)] and wireless sensor networks. Dynamic nature of the network is the main characteristic of MANET. Therefore, the SP routing problem in MANET turns into dynamic optimization problem (DOP). Here the nodes ae made aware of the environmental condition, thereby making it intelligent, which goes as the input for GA. The implementation then uses GAs with immigrants and memory schemes to solve the dynamic SP routing problem (DSPRP) in MANETS. In our paper, once the network topology changes, the optimal solutions in the new environment can be searched using the new immigrants or the useful information stored in the memory. Results shows GA with new immigrants shows better convergence result than GA with memory scheme.

## I. INTRODUCTION

MOBILE Ad hoc NETwork (MANET) is a self-organizing and self–configuring multihop wireless network, in which routing is one of the important issue that has greater impact on the performance of the network. There are mainly two types of routing protocols in MANETs, namely, topological routing and geographic routing [10].In this paper, we investigate the SP routing problem, which belongs to the topological routing. The DSPRP in MANETs is a real world dynamic optimization problem (DOP). The simplest way to address the DOPs is using evolutionary algorithms (EAs). Several approaches have been developed for EAs to address dynamic environments such as, maintaining diversity during the run via random immigrants [1], [10], increasing diversity after a change [2], using memory schemes to reuse stored useful information [3], [4], [10]. Creating awareness of the nodes to the network makes the convergence of finding the optimal path faster. Awareness requires learning of the environment. There are many techniques available for learning [6], [8] and we have used Reinforcement Learning [6] for study purpose.

In this paper, we adapt two genetic algorithms to deal with DOPs to solve the DSPRP in MANETs. Immigrants and Memory schemes and their combination into the GA are used to enhance its searching capacity for the SPs in dynamic environments. Once the network topology is changed, the optimal solutions in the new environment can be searched using the new immigrants or the useful information stored in the memory.

The rest of this paper is organized as follows. We discuss related work in Section II. The MANET network model and the DSPRP model are described in Section III. Section IV presents the design of a GA for the static SP routing problem. The GA for the DSPRP is described in Section V. The extensive experimental study and relevant analysis are presented in Section VI. Finally, Section VII concludes this paper with some discussions on the future work.

## II. RELATED WORK

Several search algorithms are investigated for solving SP routing problem. In [5], a genetic algorithm approach was presented for solving SP routing problem. Simulation studies show that the algorithm is indeed intensive to network topologies in respect of both route optimality and convergence. The quality of solution has been found to be better than other deterministic algorithms.

In [2], three modifications are applied to the Standard GA on tracking in changing environments. The modifications include: (1) the Random Immigrants mechanism, (2) increasing the mutation rate in the Standard GA, and (3) the Triggered Hyper mutation mechanism. But with the overall

This work was supported in part by the Department of Computer Science & Engineering(M Tech),NMIT, Bangalore.

Manuscript received April 21, 2011. Dr. T. R. Gopalakrishnan Nair is with the Dayananda Sagar Institutions, Bangalore, India. He is now the Director, Research Industry and Incubation Centre,DSI, Bangalore, India. (Phone: +91-80-42161766; mob: 09916586263; e-mail: trgnair@ieee.org ).

Ms.Kavitha Sooda, Asstistant Professor, Nitte Meenakshi Institute of Technology, Bnagalore-64, India, is pursuing her research program under the able guidance of Dr. T. R. Gopalakrishnan Nair, and is an Associate member of Advance Networking Research Group of Research Industry and Incubation Centre,DSI, Bangalore, India. (e-mail: kavithasooda@gmail.com).

Ms. Yashoda M B, has been pursuing M.Tech (CS&E) from NMIT, Bangalore(m.yashoda@gmail.com).

increase in mutation, the average performance deteriorates. The level of mutation may exceed the amount required, as a result, hyper mutation GA exhibit difficulties in tracking continuously changing environments. Random immigrant GA increases the probability of losing the information that may match small incremental changes in the environment.

However, these algorithms are not good choice for solving DSPRP in MANETs; in this regard we implement GA with immigrant scheme and GA with memory scheme. The elite from the previous generation and the useful information stored in memory are used to obtain the optimal solution for DSPRP in MANETs.

### III. DYNAMIC SHORTEST PATH ROUTING PROBLEM

In this section, we first present our network model and then formulate the DSPRP [10]. We consider a MANET operating within a fixed geographical region. We model it by an undirected and connected topology graph $G_0(V_0, E_0)$, where $V_0$ represents the set of wireless nodes (i.e., routers) and $E_0$ represents the set of communication links connecting two neighboring routers falling into the radio transmission range. A communication link $(i, j)$ cannot be used for packet transmission unless both node $i$ and node $j$ have a radio interface each with a common channel.

We summarize some notations that we use throughout this paper:

| | |
|---|---|
| $G(V_0,E_0)$ | Initial Manet topology graph; |
| $G(V_i, E_i)$ | MANET topology after $i^{th}$ change; |
| $s$ | Source node; |
| $d$ | Destination node; |
| $P_i(s,r)$ | Path from $s$ to $d$ in graph $G_i$; |
| $C_l$ | Cost on communication link $l$; |

The DSPRP can be informally described as follows. Initially, given a network of wireless routers, a delay upper bound, a source node, and a destination node, we wish to find a least cost loop-free path on the topology graph.

### IV. GENETIC ALGORITHM FOR SP ROUTING PROBLEM

Genetic Algorithm (GA), first introduced by John Holland in the early seventies, is the powerful stochastic algorithm based on the principles of natural selection and natural genetics, which has been quite successfully, applied in machine learning and optimization problems. To solve a problem, a GA maintains a population of individuals (also called strings or chromosomes) and probabilistically modifies the population by some genetic operators such as selection, crossover and mutation, with the intent of seeking a near optimal solution to the problem. The GA design involves several key components: genetic representation, population initialization, fitness function, selection scheme, crossover, and mutation.

#### A. Genetic representation

A routing path from source node to the destination node is encoded by a string of positive integers that represent the IDs of nodes through which the path passes. Each position of the string represents an order of a node. The gene of the first position is for the source node and the gene of the last position is for the destination node.

#### B. Population initialization

The initial population $P$ is composed of a certain number of, say $p$, chromosomes. To promote the genetic diversity, in our algorithm, the corresponding routing path is randomly generated for each chromosome in the initial population. We start to search a random path from $s$ to $d$ by randomly selecting a node $u1$ from $N(s)$, the neighborhood of $s$. Then, we randomly select a node $u2$ from $N(u1)$. This process is repeated until the destination node $d$ is reached.

#### C. Fitness function

For a given solution, evaluate its quality, which is determined by the fitness function. In our algorithm the quality of a solution is determined in terms of path cost. We aim to choose one solution with least cost among all possible solutions. Fitness value for chromosome $Ch_i$ (representing path from $s$ to $d$), denoted as $f(Ch_i)$, given by [10].

$$f(Ch_i) = \left[\sum_{l \in P(s,d)} C_l\right]^{-1} \quad (1)$$

#### D. Selection

Selection is intended to improve the average quality of the population by giving a chance to the high-quality chromosomes to get copied to the next generation. The selection of chromosome is based on the fitness value.

#### E. Crossover and Mutation

Once the individuals are selected, we are supposed to produce offspring with them. The most widely used genetic operators to drive the evolutionary process of a population in GA are crossover and mutation.

In our algorithm, since chromosomes are expressed by the path structure, we adopt the single-point crossover to exchange partial chromosomes [2]. With the crossover probability, each time we select two chromosomes $Ch_i$ and $Ch_j$ for crossover. $Ch_i$ and $Ch_j$ should possess at least one common node. After selection and crossover the population will undergo the mutation operation. With the mutation probability, each time we select one chromosome $Ch_i$ on which one gene is randomly selected as the mutation point. The mutation will replace the sub path by a new random sub path.

### V. GAS FOR DSPRP

Two applications of genetic algorithm are used to solve the dynamic shortest path routing problem[10].

- GAs with Immigrants Schemes
- GAs with Memory Schemes

#### A. GAs with Immigrants Scheme

GAs with immigrant approach uses random immigrant along with elite from previous generation to maintain diversity in the population, called *elitism-based* *immigrants*, for GAs (EIGA) to address DOPs [4], [9], [10]. Within EIGA, for each generation, after the crossover and mutation operations, the elite *E(t-1)* from previous generation *P(t-1)* is taken as the base to create immigrants in the current generation.. From elite, a set of $r_{ei} \times n$ individuals are iteratively generated by mutating $E(t-1)$ bitwise with a probability $p^i_m$, where *n* is the population size and $r_{ei}$ is the ratio of the number of elitism-based immigrants to the population size. The generated individuals then act as new immigrants and replace the worst individuals in the current population. It uses the elite from previous population to guide the immigrants toward the current environment; this way an extremely fit chromosome is never lost from our chromosome pool, which is expected to improve GA's performance in dynamic environments.

Fig. 1 shows the flowchart for EIGA. In our algorithm implementation, after crossover, mutation operation is applied on *E (t-1)* to generate new immigrants only if the mutation probability is satisfied; otherwise elite from the previous generation itself is used as new immigrant.

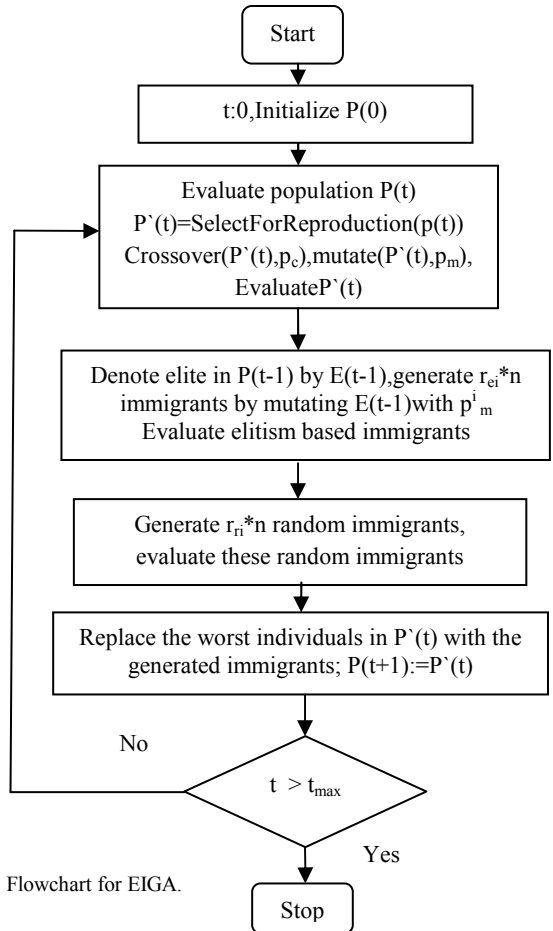

Fig. 1 Flowchart for EIGA.

### B. GAs with Memory Schemes

The memory approach works by storing best solutions from the current environment in a memory, which can be reused later in new environments. GA with memory scheme discussed in this paper is called MEGA [10]. The memory is updated in two cases, (1) each time a change in the environment is detected, and the best individual from the current environment is stored into the memory. (2) Memory is due to update, either best individual from the current generation or elite from the previous generation will replace the random points still exists in the memory; otherwise, best individual or elite will replace the most similar memory point, if it is fitter.

Fig.2 shows the flowchart for MEGA.

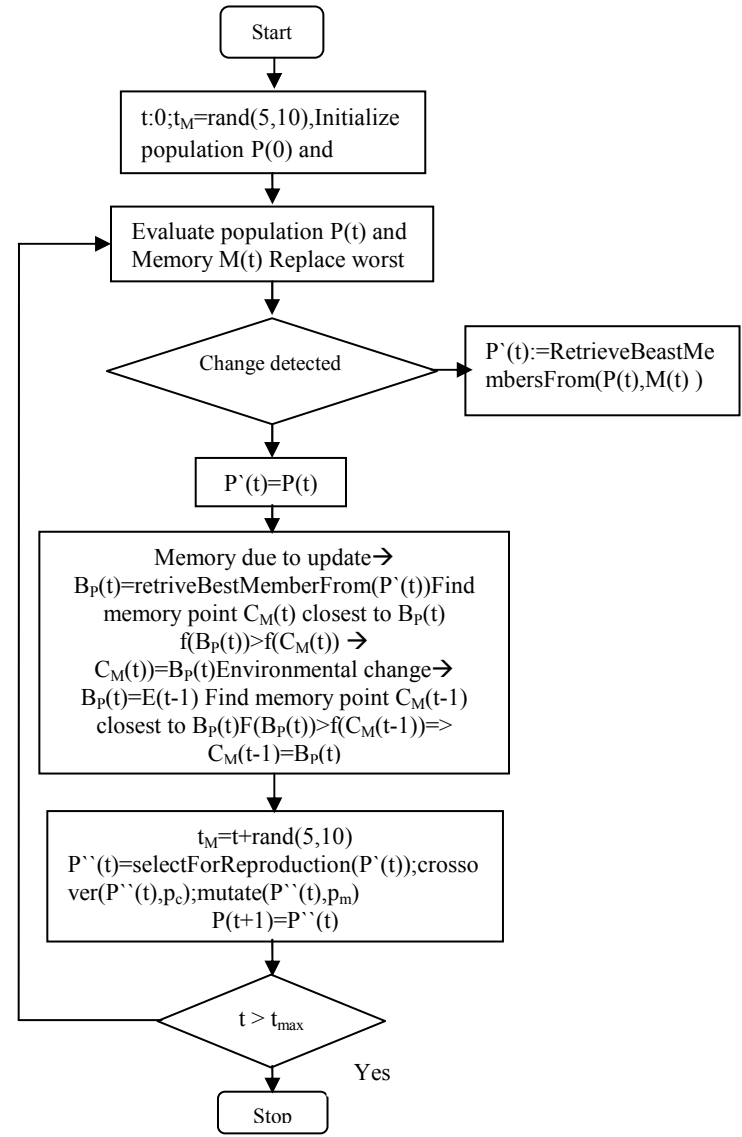

Fig. 2 Flowchart for MEGA.

## VI. EXPERIMENTAL STUDY AND ANALYSIS

### A. Network Deployment

In this paper, the initial network topology is created using random waypoint (RWP) model [7]. The random waypoint model is a commonly used model for simulations of wireless communication networks. This mobility model is a simple and straightforward stochastic model that describes the movement behavior of a mobile network node in a given system area. A node randomly chooses a destination point (waypoint) in the area and moves with constant speed on a straight line to this point. After waiting a certain pause time, it chooses a new destination and moves with constant speed towards the destination.

Fig. 3 shows the initial network topology created using RWP model. Now apply Reinforcement Learning [6] technique and obtain a concise network topology for implementing GA. Now apply EIGA, MEGA and EIGA-MEGA to de rive the results.

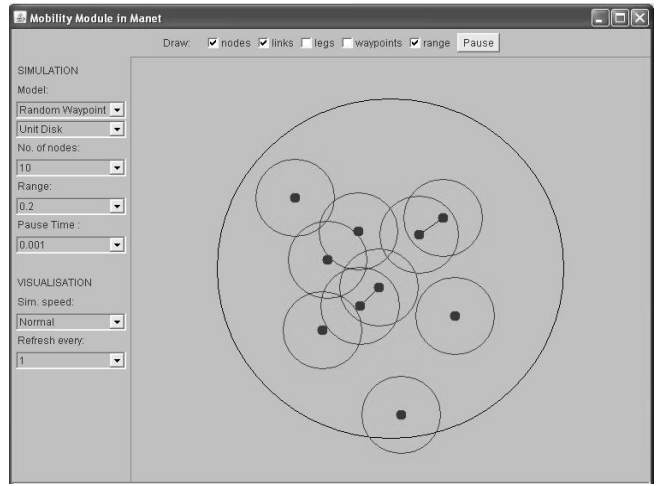

Fig. 3 Initial network topology

### B. Results

First, we consider the population size as 20 and maximum number of generations as 10. Both algorithms, EIGA and MEGA converge when the maximum numbers of generations is reached. We observe that EIGA scheme had the best fitness value as compared to the other two schemes. This fitness value of the chromosome showed the best optimal path selection, to route the packets. Fig. 4 and Fig. 5 shows the comparison result of the quality of solution for EIGA, MEGA and EIGA-MEGA schemes without knowledge base approach and with knowledge base approach respectively.

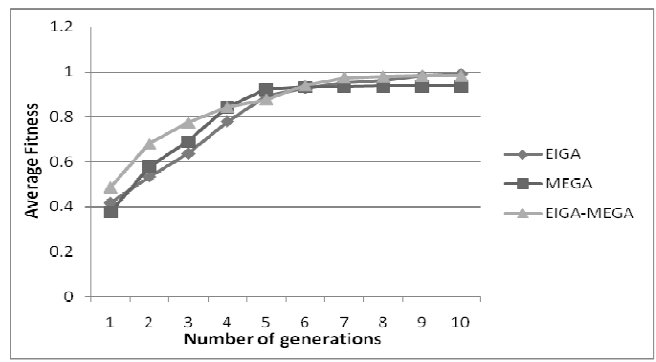

Fig. 4 Comparison result of quality of solution for EIGA, MEGA and EIGA-MEGA without knowledge base approach.

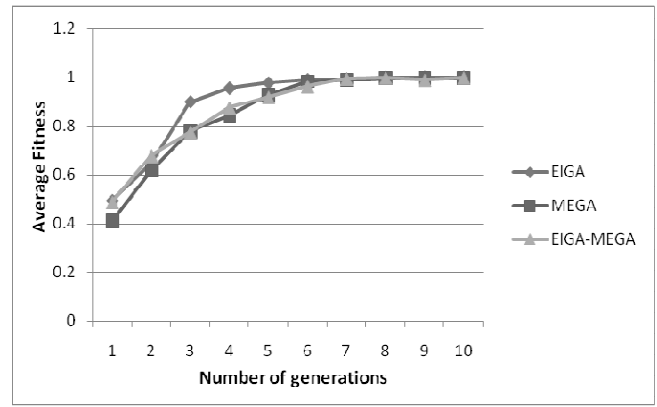

Fig. 5 Comparison result of quality of solution for EIGA, MEGA and EIGA-MEGA with knowledge base approach.

The Table I shows the comparison of EIGA, MEGA and EIGA-MEGA, depending upon the values of average fitness of given generation of chromosomes.

TABLE I

COMPARISION OF FITNESS VALUE FOR EIGA, MEGA AND EIGA-MEGA

| Generation | EIGA | MEGA | EIGA-MEGA |
|---|---|---|---|
| 1 | 0.41894 | 0.37666 | 0.4882 |
| 2 | 0.53417 | 0.57758 | 0.6828 |
| 3 | 0.63701 | 0.68911 | 0.7743 |
| 4 | 0.77900 | 0.84160 | 0.8446 |
| 5 | 0.89340 | 0.92221 | 0.8780 |
| 6 | 0.92651 | 0.93349 | 0.9408 |
| 7 | 0.95268 | 0.93413 | 0.9751 |
| 8 | 0.96157 | 0.93461 | 0.9802 |
| 9 | 0.98144 | 0.93515 | 0.9833 |
| 10 | 0.98974 | 0.93593 | 0.9839 |

The table illustrates the best possible optimal path that can be obtained using the schemes as mentioned above.

### VII. CONCLUSION AND FUTURE WORK

This paper investigates two applications of GAs for solving DSPRP in MANETs. The initial network is designed using RWP model, which is commonly used for simulations of wireless communication networks. This paper proposes elitism-based immigrant scheme for GAs in dynamic environment, where elite from the previous generation is used as the basis for creating new immigrants into the population via mutation. The memory approach works by storing useful information from the current environment for possible reuse in the dynamic environment. We observed that better results were obtained using EIGA schemes as compared to the other schemes. By implementing knowledge base approach, we observed that convergence of optimal path was 30% faster than without knowledge base approach.

We would further work on evolutionary algorithms and obtain a comparative result table to verify the performance of the GA.

ACKNOWLEDGEMENT

The authors are grateful to Dr.Jharna Majumdar, Prof. & Head, Dept of CSE (PG), Dean R&D, NMIT, Bangalore and reviewers for their thoughtful, constructive comments and suggestions.

**T.R.Gopalakrishnan Nair** holds M.Tech. (IISc, Bangalore) and Ph.D. degree in Computer Science. He has 3 decades experience in Computer Science and Engineering through research, industry and education. He has published several papers and holds patents in multi domains. He has won the PARAM Award for technology innovation. Currently he is the Director of Research and Industry in Dayananda Sagar Institutions, Bangalore, India.

**Kavitha Sooda** holds M.Tech in Computer Scinece and Engineering. She has nine years of teaching experience and pursuing her Ph. D from JNTU, Hyderbad. Her interest includes rouitng techniques, QoS application, cognitive networks and genetic alrothims. Currently she works as Asstiant Profressor at Nitte Meenakshi Institute of Technology, Bangalore, India.

**Ms Yashoda M B** pursuing M.Tech degree in Computer Science and Engineering at NMIT, Yelehanka, Bangalore. Her work includes the in-depth study of evolutionary algorithm, routing techniques and analysis of intelligent routing approaches.